\documentclass{article}

\usepackage{subcaption}
\usepackage[colorinlistoftodos,prependcaption,textsize=small]{todonotes}

\usepackage[preprint]{neurips_2025}

\usepackage[utf8]{inputenc} 
\usepackage[T1]{fontenc}    
\usepackage{hyperref}       
\usepackage{url}            
\usepackage{booktabs}       
\usepackage{amsfonts}       
\usepackage{nicefrac}       
\usepackage{microtype}      
\usepackage{xcolor}         
\usepackage{amsmath}
\usepackage{graphicx}

\title{LINKER: \underline{L}earning \underline{In}teractions Between Functional Groups and Residues With Chemical \underline{K}nowledge-\underline{E}nhanced \underline{R}easoning and Explainability}

\author{
  Phuc Pham$^{1,2}$ \quad
  Viet Thanh Duy Nguyen$^{1}$ \quad
  Truong-Son Hy$^{1}$\thanks{Corresponding author: \texttt{thy@uab.edu}} \\
  \\
  $^{1}$Department of Computer Science, University of Alabama at Birmingham, Alabama, United States \\
  $^{2}$Faculty of Computer Science and Engineering, Ho Chi Minh University of Technology, HCM, Viet Nam \\
}

\begin{document}

\maketitle

\begin{abstract}
Accurate identification of interactions between protein residues and ligand functional groups is essential to understand molecular recognition and guide rational drug design. Existing deep learning approaches for protein-ligand interpretability often rely on 3D structural input or use distance-based contact labels, limiting both their applicability and biological relevance. We introduce LINKER, the first sequence-based model to predict residue-functional group interactions in terms of biologically defined interaction types, using only protein sequences and the ligand SMILES as input. LINKER is trained with structure-supervised attention, where interaction labels are derived from 3D protein-ligand complexes via functional group-based motif extraction. By abstracting ligand structures into functional groups, the model focuses on chemically meaningful substructures while predicting interaction types rather than mere spatial proximity. Crucially, LINKER requires only sequence-level input at inference time, enabling large-scale application in settings where structural data is unavailable. Experiments on the LP-PDBBind benchmark demonstrate that structure-informed supervision over functional group abstractions yields interaction predictions closely aligned with ground-truth biochemical annotations.
\end{abstract}

\section{Introduction}

Protein-ligand interactions underpin virtually all processes in chemical biology and pharmacology. Understanding how a ligand engages with its target protein, particularly which functional groups form which types of interactions with specific residues, is critical to elucidating molecular recognition mechanisms, rationalizing structure-activity relationships, and guiding lead optimization in drug discovery. In existing work, obtaining detailed protein-ligand interaction maps almost always requires access to a 3D complex structure. However, such complexes are not always experimentally available and, when they are missing, researchers typically resort to molecular docking between a protein structure and a ligand using simulation tools such as AutoDock Vina \cite{trott2010autodock}, followed by post-processing with tools such as PLIP \cite{salentin2015plip} to extract interaction types. This process can be computationally expensive and time-consuming, particularly in the case of blind docking, where the binding pocket is unknown and the search space is large.

To the best of our knowledge, there is no prior method that predicts biologically defined residue-functional group interactions directly from sequence-level input. Some sequence-based binding affinity prediction models incorporate structure-supervised attention and generate residue-ligand interaction maps as an explanatory by-product. Because interaction prediction is treated only as an auxiliary objective in these models, its accuracy is often limited. Other sequence-based approaches aim to predict the residue pocket directly from the sequence, but they either treat the ligand as a flat list of atoms or supervise attention using only interatomic distance labels. Both cases lack functional group abstractions and biologically defined interaction types, which constrains their chemical interpretability.

In this work, we present LINKER, the first sequence-based framework for directly predicting residue-functional group interaction types using only protein sequences and ligand SMILES as input. LINKER is trained with \emph{structure-supervised attention}, where supervision comes from interaction labels extracted by PLIP from experimentally determined protein-ligand complexes, after the ligands are abstracted into chemically meaningful functional groups. This design enables the model to focus on the substructures that drive interactions and to predict biologically defined interaction types, such as hydrogen bonds, $\pi$-stacking, or salt bridges, rather than raw interatomic distances. Crucially, LINKER operates entirely at the sequence level during inference, allowing large-scale application to protein-ligand pairs lacking experimental structures while retaining chemical interpretability aligned with medicinal chemistry reasoning.

\section{Problem Formulation}
\label{sec:problem_formulation}

We cast residue--functional group interaction prediction as a multi-label classification task. 
Given a protein--ligand pair, the objective is to estimate, for every protein residue and ligand functional group, independent probabilities over seven biologically defined interaction types: hydrogen bonds, hydrophobic contacts, $\pi$--stacking, $\pi$--cation interactions, salt bridges, water bridges, and halogen bonds. 
Formally, let $\mathbf{D}$ denote the ligand SMILES sequence, decomposed into $F$ functional groups, and let $\mathbf{T}$ denote the protein amino acid sequence, comprising $R$ residues. 
Our model, LINKER, learns the following mapping:
\[
f_{\text{LINKER}}: (\mathbf{D}, \mathbf{T}) \;\rightarrow\; \mathbf{P} \in [0,1]^{R \times F \times 7},
\]
where $\mathbf{P}_{r,f,k}$ denotes the likelihood that residue $r$ and functional group $f$ are involved in the interaction type $k$, with $k = 1, \dots, 7$. Unlike contact maps that only indicate spatial proximity—where residues may be close but not engaged in any biochemical interaction—LINKER provides direct supervision over interaction types, enabling chemically grounded reasoning and improved interpretability.

\section{Method}
\begin{figure}[h]
    \centering
    \includegraphics[width=\linewidth]{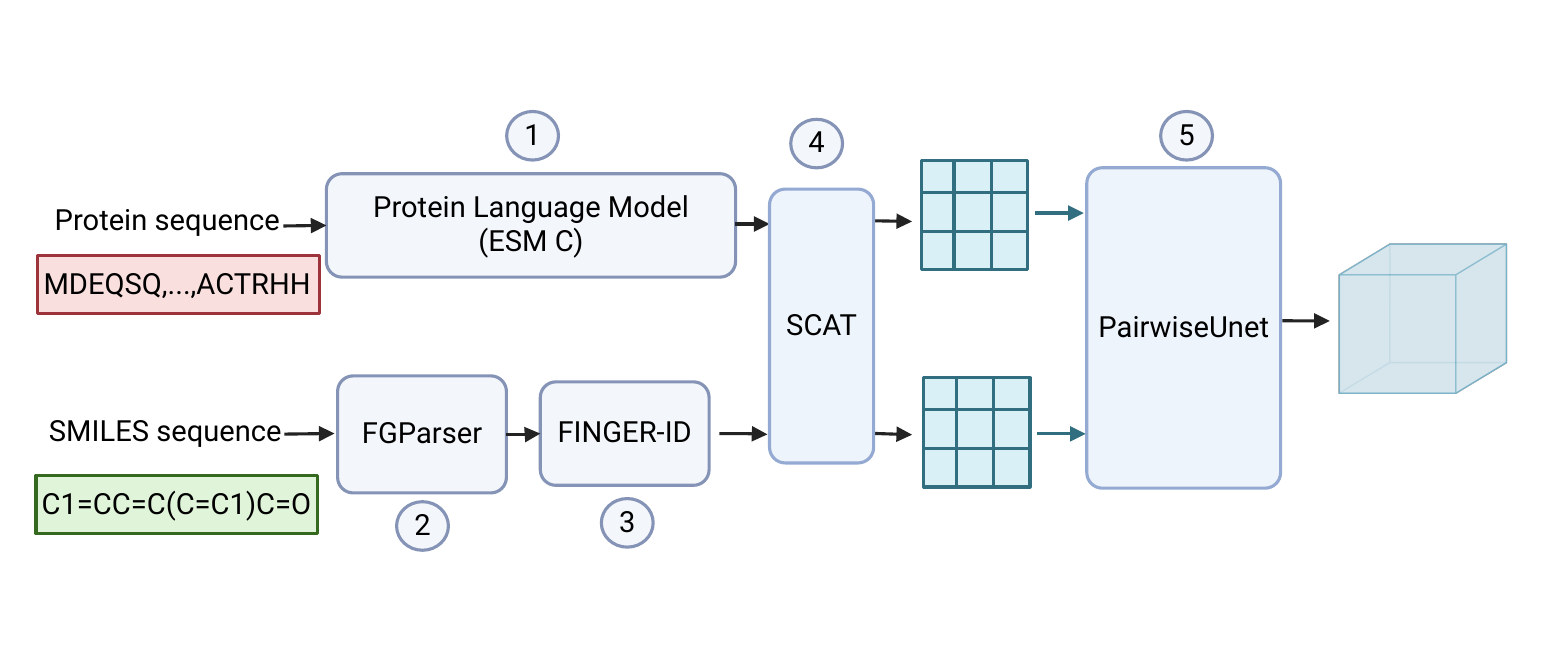}
    \caption{
        Overview of the LINKER framework for predicting biologically defined residue--functional group interaction types from protein sequences and ligand SMILES. 
        The core components include: 
        (1) Protein Language Model (ESM C), which encodes protein sequences into contextual residue embeddings; 
        (2) Functional Group Parser (FGParser), which extracts functional groups from ligand SMILES and produces an atom--group representation matrix; 
        (3) Functional Group and Positional Embedding for Molecular Identification (FINGER-ID), which generates context-aware functional group embeddings with positional and structural information; 
        (4) Self \& Cross Attention Transformer (SCAT), which integrates intra- and inter-molecular context between residues and functional groups; and 
        (5) PairwiseUNet, which predicts interaction probability maps over all residue--functional group pairs and interaction types.
    }
    \label{fig:framework}
\end{figure}

We propose LINKER, a sequence-based framework for predicting biologically defined residue-functional group (FG) interaction types from protein sequences and ligand SMILES. As illustrated in Figure ~\ref{fig:framework}, LINKER consists of two modality-specific encoding branches, a cross-attention integration module, and a pairwise interaction predictor. A detailed description of each module can be found in Appendix~\ref{sec:detailed_methodology}.

\paragraph{Protein branch}
The protein sequence is encoded using the ESM C~\cite{esm2024cambrian} protein language model, which is optimized to capture biologically meaningful representations of proteins. Unlike the ESM-3 \cite{hayes2024simulating} generative models focused on controllable sequence generation, \textit{ESM C} specializes in representation learning and supports multi-chain inputs, making it more versatile than ESM-2 \cite{lin2022language} for modeling complex protein-ligand interactions without requiring structural data.

\paragraph{Ligand branch}
Inspired by the development of multiscale models for complex chemical systems~\cite{karplus2014development}, we adopt a higher-level molecular representation based on functional groups rather than individual atoms. This abstraction aligns more naturally with protein-ligand interactions, which are typically governed by chemically meaningful substructures in the ligand and complementary residues in the protein. Traditional fingerprints such as the Morgan fingerprint~\cite{morgan1965generation}, also known as the extended-connectivity fingerprint (ECFP4)~\cite{rogers2010extended}, do capture local substructures, but they represent atom-centered environments without global context. As a result, the generated fragments often correspond to partial motifs that are difficult to consolidate into chemically interpretable functional groups, and multiple instances of the same group within a molecule cannot be distinguished. The lack of positional and semantic interpretability limits their explanatory power for binding mechanisms, thereby motivating our use of explicit functional group representations, which provide both chemical interpretability and biological relevance.

To address these limitations, we introduce two complementary modules. The \textit{Functional Group Parser (FGParser)} deterministically maps each atom in a ligand SMILES to a functional group and interpolates the resulting group features back onto atoms to produce an \(N_{\text{atom}}\!\times\!F\) atom-group representation matrix. Built upon this mapping, the \textit{Functional Group and Positional Embedding for Molecular Identification (FINGER-ID)} learns multiscale representations by integrating local atomic neighborhood information with learnable functional-group embeddings that capture intrinsic chemical properties, while augmenting each group embedding with positional encodings and graph-level context derived from a Graph Convolutional Network (GCN)~\cite{kipf2017semi}. Our hierarchical architecture preserves atom–group correspondence and integrates signals across multiple scales, yielding position-aware and chemically interpretable representations that improve binding and molecular-property prediction.

\paragraph{Interaction context integration}
To model both intra- and inter-molecular dependencies, we introduce the \textit{Self \& Cross Attention Transformer (SCAT)} module. SCAT first applies self-attention~\cite{vaswani2017attention} separately to the protein residue embeddings and ligand functional group embeddings to capture context within each modality. It then performs bidirectional cross-attention, where residue embeddings attend to functional group embeddings and vice versa, enabling the model to focus on residue–FG pairs with high interaction likelihood. The output is updated residue and functional group representations that are enriched with both local sequence/chemical context and cross-molecular interaction signals.

\paragraph{Pairwise interaction prediction}
The enriched protein and ligand representations produced by SCAT are integrated into a pairwise feature tensor, where each position corresponds to a residue–functional group pair. This tensor is then processed by \textit{PairwiseUnet}, a customized architecture inspired by the 2D U-Net \cite{ronneberger2015u}, designed to capture both local interaction motifs and global structural patterns within the interaction map. By casting interaction prediction as an image-like task, PairwiseUnet preserves fine-grained residue-group alignments while simultaneously modeling long-range dependencies, enabling the discovery of reusable biochemical motifs such as hydrogen bonds, hydrophobic contacts, and $\pi$-stacking. The network outputs a probability distribution over biologically defined interaction types for each residue-functional group pair, yielding binding maps that are both mechanistically meaningful and directly accessible from sequence-level input.

\section{Experiments}
In this section, we outline the experimental setup used to evaluate LINKER, including dataset preparation, baseline comparisons, and performance results on the held-out test set. For additional implementation details, please refer to Appendix~\ref{sec:appendix_implementation_details}.

\subsection{Dataset}
We evaluated LINKER on protein-ligand complexes from the PDBBind dataset \cite{liu2015pdb}, which provides high-quality structural data for a wide variety of binding interactions. Although PDBBind is commonly used for binding affinity prediction, in our work, it serves as a source of experimentally resolved 3D complexes from which we derive residue-functional group interaction labels. Ground-truth labels are obtained by applying PLIP \cite{salentin2015plip} to each complex after the ligands are decomposed into functional groups.

To ensure a realistic and unbiased evaluation, we adopt the Leak-Proof PDBBind split \cite{li2024leak}, which eliminates structural redundancy between training and test sets by clustering protein-ligand complexes based on binding site similarity. This is particularly important for LINKER, as the supervision signal for attention is derived from 3D structural motifs. Preventing structural overlap ensures that LINKER cannot rely on memorized interaction patterns and must instead learn generalizable, interpretable mappings between sequence-level representations and biochemical interaction types.

\subsection{Baselines}

Most prior models generate protein-ligand interaction maps as a secondary outcome of binding affinity prediction, typically using attention visualization or backpropagation rather than direct supervision \cite{monteiro2022dtitr, mcnutt2024sprint, hu2025multi}. These approaches often require 3D protein structures at inference time and may yield attention weights that do not reliably correspond to true biochemical interactions.

ArkDTA \cite{gim2023arkdta} stands out as the only sequence-based method that explicitly supervises cross-modal attention using residue-level interaction labels derived from 3D complex structures. By aligning attention weights between protein residues and ligand substructure tokens with non-covalent interaction (NCI) annotations, ArkDTA enables structure-free interpretability during inference, and thus serves as our primary baseline.

However, ArkDTA's supervision is limited to coarse residue-level labels and relies on Morgan fingerprint-based substructures, which lack chemical interpretability. In contrast, our method predicts fine-grained residue-functional group interactions using chemically meaningful groups extracted by PyCheckMol. To ensure a fair comparison despite these differences, we evaluated both models on the shared task of predicting interaction residues under aligned supervision settings.

\subsection{Results}
\subsubsection{Residue Interaction Prediction}

We evaluated the predictive performance of LINKER in identifying interaction residues, i.e., protein residues involved in ligand binding, using hard binary labels derived from PLIP annotations. To enable a fair comparison with ArkDTA, which is supervised only at the residue level, we aggregated our finer-grained residue-functional group interaction labels into residue-level binary indicators, where a residue is marked as positive if it interacts with at least one ligand atom. We then compare LINKER with ArkDTA using precision-recall (PR) and receiver operating characteristic (ROC) curves computed over all protein residues. These metrics, respectively, highlight robustness under severe class imbalance and overall discriminative ability in distinguishing interacting from non-interacting residues.


To enable this comparison, we derive residue-level interaction scores from LINKER's output tensor defined in Section~\ref{sec:problem_formulation}, where the model predicts a tensor $\mathbf{P} \in [0,1]^{R \times F \times 7}$ representing interaction probabilities between $R$ protein residues and $F$ ligand functional groups in seven biologically defined interaction types. 
We first aggregate over functional groups by taking the maximum along the $F$-dimension:
\[
\mathbf{U}_{r,k} = \max_{1 \leq f \leq F} \mathbf{P}_{r,f,k}, \quad \text{for } r = 1, \dots, R \text{ and } k = 1, \dots, 7,
\]
resulting in a residue-wise interaction score matrix $\mathbf{U} \in [0,1]^{R \times 7}$. 
Next, we aggregate over interaction types:
\[
\mathbf{y}_r = \max_{1 \leq k \leq 7} \mathbf{U}_{r,k}, \quad \text{for } r = 1, \dots, R,
\]
which produces a final prediction vector at the residue level $\mathbf{y} \in [0,1]^R$. 
This transformation produces a single interaction confidence score per residue, enabling a fair comparison with ArkDTA, which is supervised at the residue level. 
Importantly, this reduction preserves the fine-grained interaction information learned by LINKER while aligning it with the coarser evaluation setting.

\begin{figure*}[h!]
    \centering
    \begin{subfigure}[t]{0.49\linewidth}
        \centering
        \includegraphics[width=\linewidth]{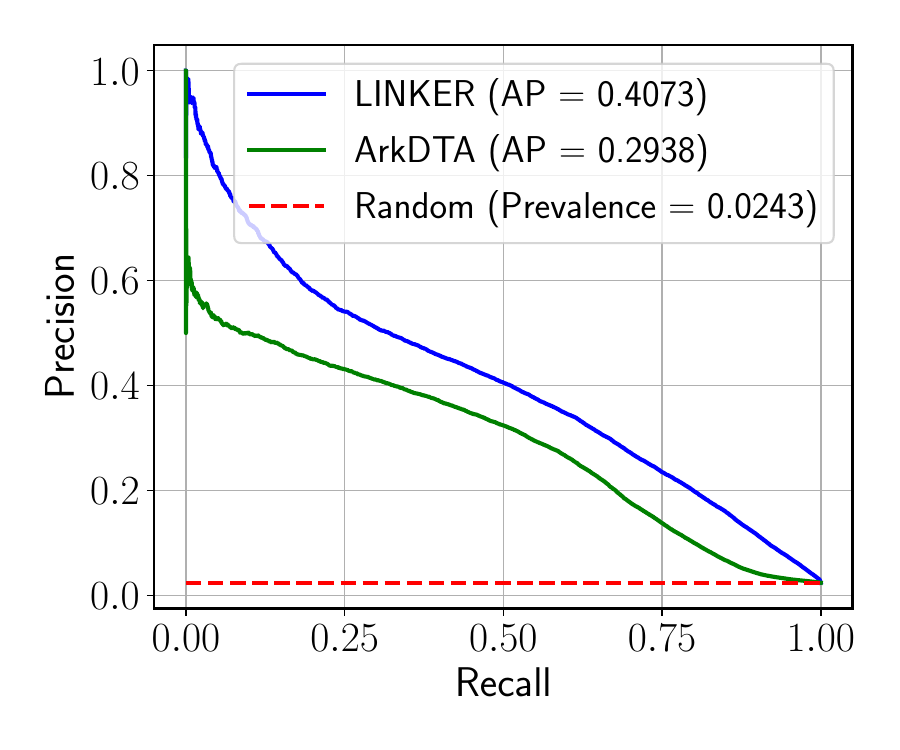}
        \label{fig:pr}
    \end{subfigure}
    \hfill
    \begin{subfigure}[t]{0.49\linewidth}
        \centering
        \includegraphics[width=\linewidth]{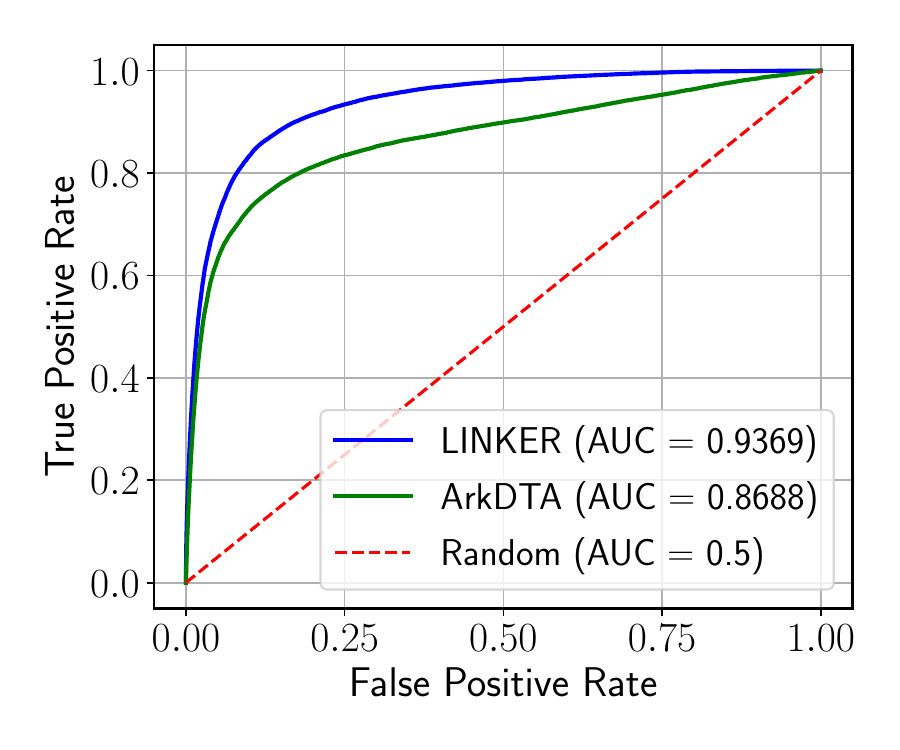}
        \label{fig:roc}
    \end{subfigure}
    \vspace{-20pt}
    \caption{Residue-level interaction prediction. LINKER consistently surpasses ArkDTA in both PR \textit{(left)} and ROC \textit{(right)}. The PR analysis emphasizes robustness under class imbalance, while the ROC highlights stronger overall discrimination.}
    \label{fig:pr-roc}
\end{figure*}

As shown in Figure~\ref{fig:pr-roc}, LINKER achieves a substantially higher area under the precision-recall curve (AP = 0.4073) compared to ArkDTA (AP = 0.2938), while the prevalence of positives (random baseline) is only 0.0243. This demonstrates LINKER's improved sensitivity and precision in detecting true interaction residues under severe class imbalance. Similarly, LINKER achieves a higher ROC AUC score (AUC = 0.9369) versus ArkDTA (AUC = 0.8688), with a random baseline of 0.5, confirming its overall superior classification performance. These results underscore the effectiveness of our interaction supervision strategy and LINKER's ability to extract biologically meaningful binding signals from sequence-only inputs.

\subsubsection{Residue Interaction Prediction with Soft Labels}

Although PLIP-derived interaction labels provide reliable supervision, they are inherently binary and rely on strict geometric cutoffs, which can introduce boundary artifacts. In reality, ligand binding is often mediated by clusters of neighboring residues, where residues adjacent in sequence and spatially proximal may all contribute to recognition, even if only a subset is captured by PLIP's criteria. To account for this biological continuity and reduce sensitivity to cutoff-induced noise, we introduce Gaussian kernel smoothing along the residue dimension. Concretely, let us donote 
\[
y_{\text{hard}}[i] = \mathbb{I}\!\left[\sum_{j} \mathbf{Y}_{i,j} > 0 \right]
\] 
as the binary residue-level label from PLIP. The smoothed label for the residue \( i \) is then defined as
\[
y_{\text{smooth}}[i] =
\begin{cases}
\displaystyle \max_{c \in H} \exp\!\left(-\frac{(i - c)^2}{2\sigma^2}\right), & H \neq \emptyset, \\
0, & H = \emptyset,
\end{cases}
\quad H = \{\,c \mid y_{\text{hard}}[c] = 1\,\}.
\]
Here, PLIP-identified residues serve as anchor points with full confidence, while their immediate neighbors receive smoothly decaying support. Rather than altering the underlying ground truth, this smoothing acts as a biologically motivated regularization that reflects the clustered nature of binding sites. In doing so, it mitigates cutoff-induced artifacts and produces graded signals that enable a more flexible and faithful evaluation of interpretability.

\begin{figure}[h]
    \centering
    \includegraphics[width=\linewidth]{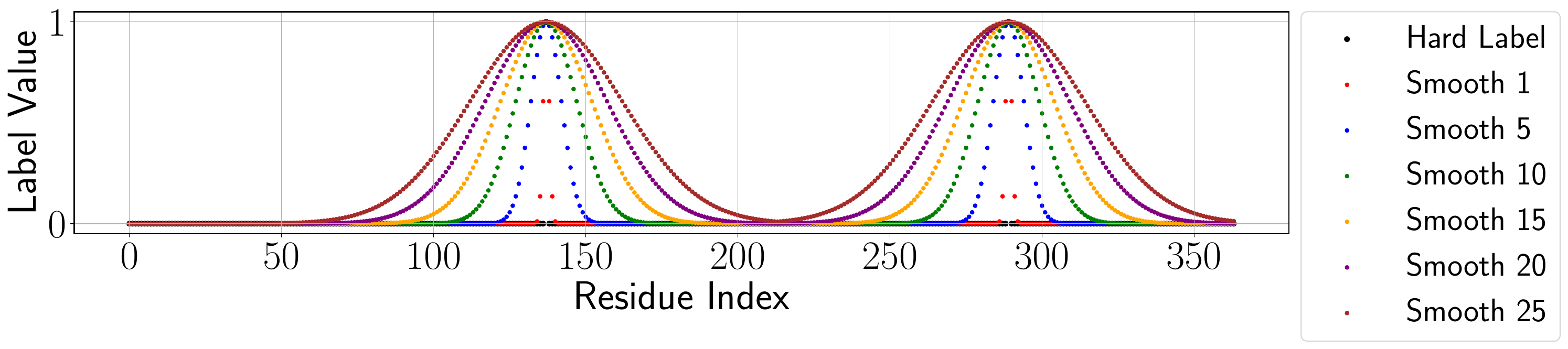}
    \caption{Visualization of residue-level interaction labels with Gaussian smoothing at different kernel widths. Wider kernels yield smoother label distributions, providing soft supervision signals around interacting residues.}
    \label{fig:smooth_label_vis}
\end{figure}

 Figure~\ref{fig:smooth_label_vis} illustrates the effect of smoothing with various standard deviations, showing how the binary interaction labels become progressively more diffuse as the smoothing parameter increases.

\begin{figure}[h]
    \centering
    \includegraphics[width=\linewidth]{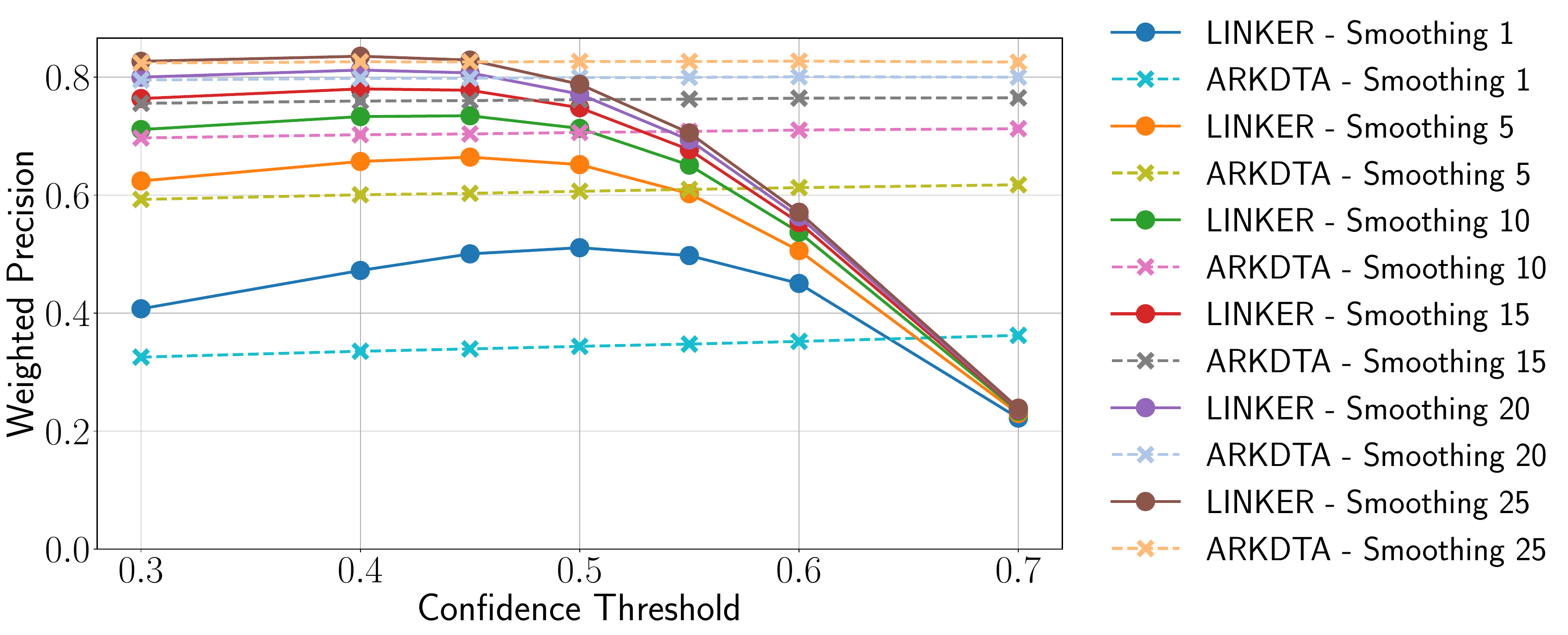}
    \caption{Weighted precision of LINKER and ArkDTA across various levels of label smoothing and confidence thresholds.}
    \label{fig:smooth_precision}
\end{figure}

To evaluate model performance under this soft supervision regime, we computed the weighted precision of predicted residue-functional group interactions at different confidence thresholds, comparing LINKER and ArkDTA across smoothing levels. As shown in Figure~\ref{fig:smooth_precision}, LINKER consistently outperforms ArkDTA on all smoothing scales and thresholds. Notably, LINKER achieves the highest weighted precision with a moderate smoothing parameter, indicating its ability not only to localize precise interaction sites but also to capture a broader biochemical context. In contrast, ArkDTA shows limited sensitivity to the smoothing factor, with relatively flat performance curves.

\subsubsection{Residue-Functional Group Interaction Prediction}

We evaluated LINKER's ability to predict fine-grained interactions between protein residues and ligand functional groups, a novel task that, to our knowledge, has not been directly addressed in prior work. To evaluate LINKER’s performance on this new task, we compare its predictions against ground truth labels derived from PLIP and report both quantitative metrics and qualitative visualizations.

\begin{figure*}[t]
    \centering
    \begin{subfigure}[t]{0.49\linewidth}
        \centering
        \includegraphics[width=\linewidth]{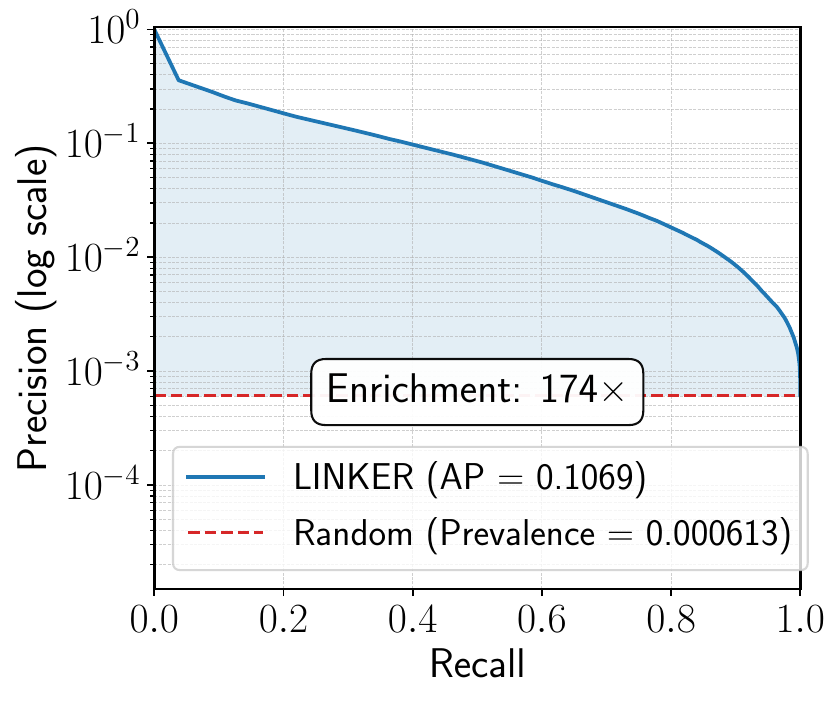}
        \caption{PR curve with enrichment over the prevalence baseline (dashed).}
        \label{fig:pr_fg_residue}
    \end{subfigure}\hfill
    \begin{subfigure}[t]{0.49\linewidth}
        \centering
        \includegraphics[width=\linewidth]{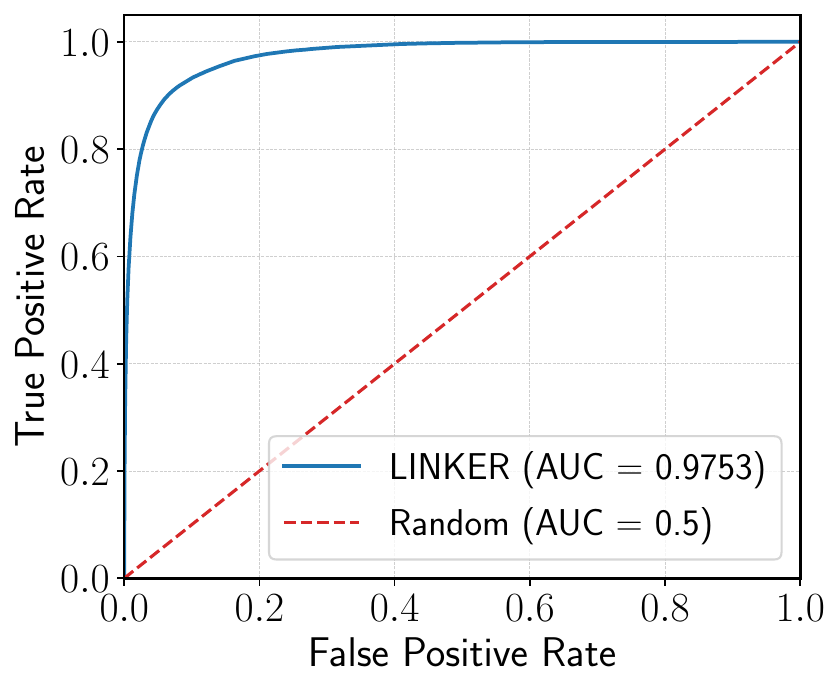}
        \caption{ROC curve showing strong discrimination between interacting and non-interacting pairs.}
        \label{fig:roc_fg_residue}
    \end{subfigure}

    \caption{Residue-Functional Group interaction prediction. LINKER delivers markedly higher enrichment at low recall and strong overall discrimination compared to a random baseline.}
    \label{fig:pr-roc-fg-residue}
\end{figure*}

\textbf{Quantitative Evaluation.} Figure~\ref{fig:pr-roc-fg-residue} reports LINKER’s performance on residue–functional-group prediction. The PR curve (Fig.~\ref{fig:pr_fg_residue}) is shown against a random baseline (prevalence = 0.000613). Model strength is summarized by fold enrichment—defined as precision relative to prevalence—and all numerical evaluations, including fold enrichment and average precision, are consistently performed on the linear scale. Because prevalence is vanishingly small and raw precision values cluster near zero, we plot precision on a logarithmic axis for visualization only. Under this setting, LINKER achieves up to 174× enrichment over random at low recall, demonstrating substantial recovery of true residue–functional-group interactions despite extreme class sparsity (i.e., the positive class is exceedingly rare relative to negatives).

The ROC curve in Figure~\ref{fig:roc_fg_residue} shows the trade-off between the true positive rate and false positive rate. LINKER attains an AUC of $0.9753$, indicating excellent discrimination between interacting and non-interacting residue–functional group pairs. Together, the PR and ROC analyses confirm that LINKER reliably captures rare interaction signals despite severe class imbalance.


\begin{figure}[h]
\centering
\includegraphics[width=\linewidth]{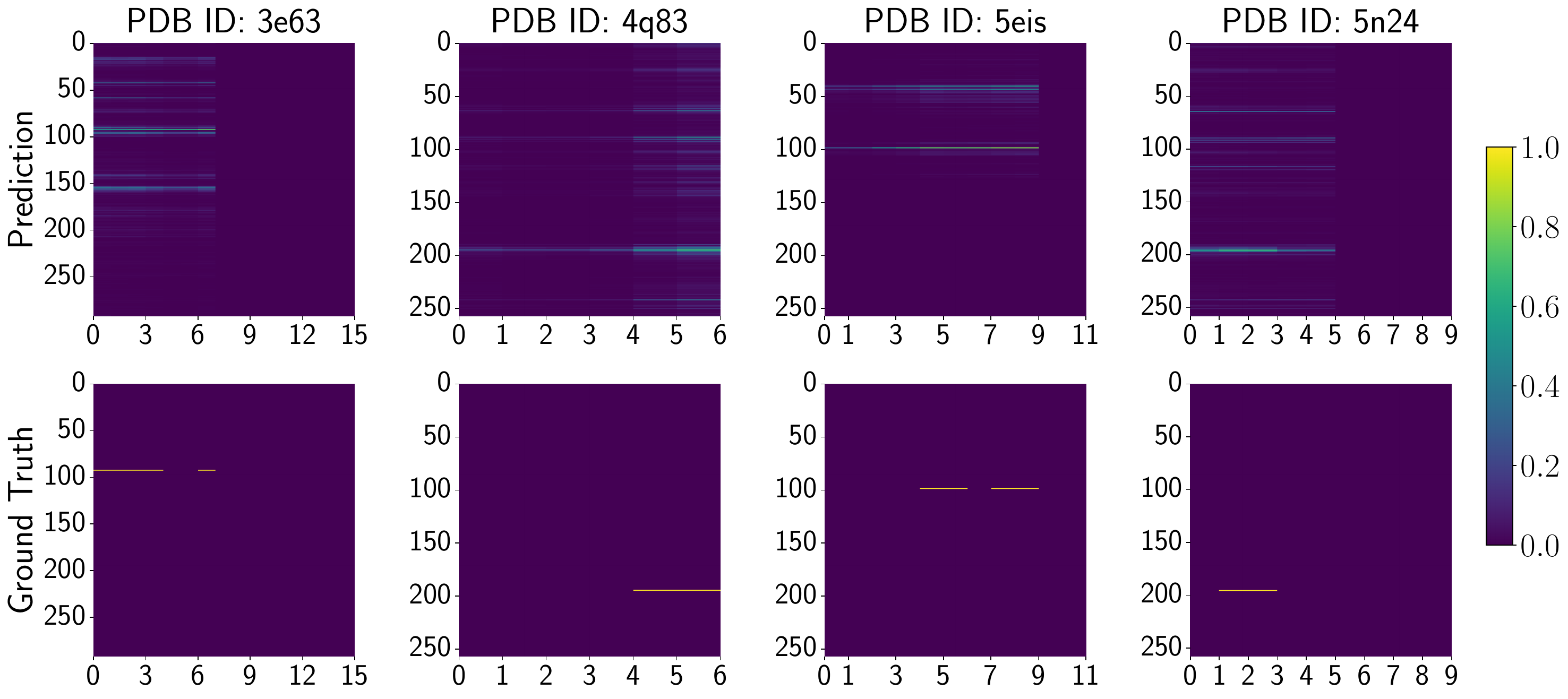}
\caption{Comparison between LINKER predictions and PLIP ground truth for hydrophobic contacts across four protein structures. the x-axis shows functional group indices derived from the atom-group matrix of the FGParser module \ref{sec:fgparser}, and the y-axis corresponds to residue indices.}
\label{fig:Linker-pdb}
\end{figure}

\paragraph{Qualitative Evaluation.} To further assess the interpretability of LINKER's predictions, we visualize the predicted residue-functional group interaction maps across four representative protein-ligand complexes in Figure~\ref{fig:Linker-pdb}. For each example, we compare the predicted interaction probabilities with the corresponding PLIP-derived ground truth. LINKER consistently highlights spatially localized interaction regions that align well with the annotated contacts, with the confidence scores highest for residue-group pairs in close spatial proximity to the ground truth. This shows the model's ability to infer chemically meaningful residue-group associations from sequence-level input while calibrating its predictions according to structural plausibility. Notably, even in complexes with sparse ground-truth labels, LINKER maintains high specificity by suppressing spurious interactions, illustrating its robust generalization. These examples confirm that LINKER produces interpretable and biologically grounded interaction maps across diverse structural contexts.

\subsection{Transferability of LINKER Representations to Binding Affinity Prediction}

To assess the generalization of our protein–ligand representations beyond interaction type prediction, we evaluated their performance on a downstream binding affinity regression task. This experiment validates whether LINKER captures biologically meaningful and transferable information about binding strength. Architectural specifications of the Binding Affinity Predictor are provided in Appendix~\ref{app:affinity_predictor}, and implementation details in Appendix~\ref{sec:appendix_implementation_details}.

We perform this evaluation on the Leak-Proof PDBBind benchmark, a non-redundant, high-quality dataset designed to test model generalization under strict sequence and structural similarity constraints. To contextualize LINKER's performance, we compare it against a range of existing methods that also report results on this benchmark, spanning sequence-based, structure-based, and multimodal approaches. This enables a fair assessment of the transferability of LINKER’s learned representations to a challenging real-world prediction task.

As shown in Table~\ref{tab:affinity_results}, LINKER achieves a test RMSE of 1.47, which is competitive with or better than several state-of-the-art methods specifically designed for the prediction of binding affinity. Notably, despite not being explicitly trained for this task, LINKER outperforms traditional docking-based approaches such as AutoDock Vina and learning-based baselines such as DeepDTA and IGN. Although its train and validation errors are slightly higher than those of MPRL and ArkDTA, both of which are affinity-focused models, LINKER achieves equivalent generalization performance on the test set, highlighting the transferability and biological relevance of its learned representations. This result confirms that LINKER captures fundamental interaction patterns that are predictive of binding strength, demonstrating strong potential for use in broader drug discovery pipelines.

\begin{table}[ht]
\centering
\caption{Comparison of RMSE on the Leak-Proof PDBBind benchmark for binding affinity prediction.}
\vskip 0.15in
\begin{tabular}{lccc}
\toprule
Model & Train & Validation & Test \\
\midrule
AutoDock Vina \cite{trott2010autodock} &2.42&2.29&2.56 \\
InteractionGraphNet (IGN) \cite{jiang2021interactiongraphnet} &1.65&2.00&2.16 \\
Random Forest (RF)-Score \cite{ballester2010machine} &0.68&2.14&2.10 \\
{DeepDTA \cite{ozturk2018deepdta}} &1.41&2.07&2.29 \\
MPRL \cite{nguyen2024multimodal} & \textbf{0.48} & 1.47 & 1.55 \\
ArkDTA \cite{gim2023arkdta} & 1.18 & \textbf{1.47} & 1.48 \\
\midrule
LINKER (Our) & 1.38 & 1.53 & \textbf{1.47} \\
\bottomrule
\end{tabular}
\label{tab:affinity_results}
\end{table}

\section{Conclusion}

In this work, we introduced LINKER, a sequence-based model for predicting biologically defined residue-functional group interaction types directly from protein sequences and the ligand SMILES. By leveraging structure-supervised attention and motif extraction based on functional group, LINKER captures chemically meaningful substructures and predicts interaction types beyond simple spatial contacts. Operating entirely at the sequence level during inference, it enables large-scale application in scenarios lacking 3D structural data. Our experiments on the LP-PDBBind benchmark confirm that structure-informed supervision over functional group abstractions produces predictions that are highly consistent with ground-truth biochemical annotations, highlighting LINKER's potential as a practical and interpretable tool for molecular recognition and rational drug design.

In future work, we plan to evaluate LINKER on a broader range of protein-ligand interaction datasets to assess its generalization and robustness. We also aim to extend the framework to other types of biomolecular interactions, such as protein-protein, antibody-antigen, and protein-RNA binding, because the same core idea can be easily adapted to these tasks, broadening its applicability across diverse areas of structural biology and therapeutic design.

\newpage
\bibliographystyle{plain}
\bibliography{main}

\newpage
\appendix
\section{Detailed Methodology}
\label{sec:detailed_methodology}

\subsection{Protein Language Model (ESM C)} \label{sec:esmc}

Let the input protein target be a FASTA sequence 
\(
\mathbf{T} = (t_1,\ldots,t_R), \; t_r \in \mathcal{A},
\) 
where \(\mathcal{A}\) denotes the amino acid alphabet and \(R\) is the sequence length.  
The sequence is tokenized and passed through the ESM C encoder, which outputs residue-level embeddings:
\[
\mathbf{H}_p\ = f_\theta(\mathbf{T}):\; 
\mathcal{A}^R \;\longrightarrow\; \mathbb{R}^{R \times D}, 
\qquad 
\mathbf{H}_p\ = 
\begin{bmatrix}
\mathbf{h}_1^\top \\ 
\vdots \\ 
\mathbf{h}_R^\top
\end{bmatrix} \in \mathbb{R}^{R \times D}.
\]

The ESM C module provides contextualized residue embeddings, where each
\(\mathbf{h}_r\) captures both the local biochemical properties of residue \(t_r\) and its long-range dependencies within the full protein sequence.  
These embeddings serve as the protein representation for downstream modules, enabling our framework to take advantage of rich structural and evolutionary information learned from large-scale protein corpora, without requiring explicit 3D structures. In particular, the residue-level features \(\mathbf{H}_p\) are subsequently paired with ligand representations to facilitate accurate interaction and binding affinity prediction. In our setting, we use the ESM C (300M) variant \cite{esm2024cambrian} with hidden size \(D=960\), resulting in an output matrix \(\mathbf{H}_p\in\mathbb{R}^{R\times 960}\).

\subsection{Functional Group Parser (FGParser)} \label{sec:fgparser}
\begin{figure}[h]
\centering
\includegraphics[width=\linewidth]{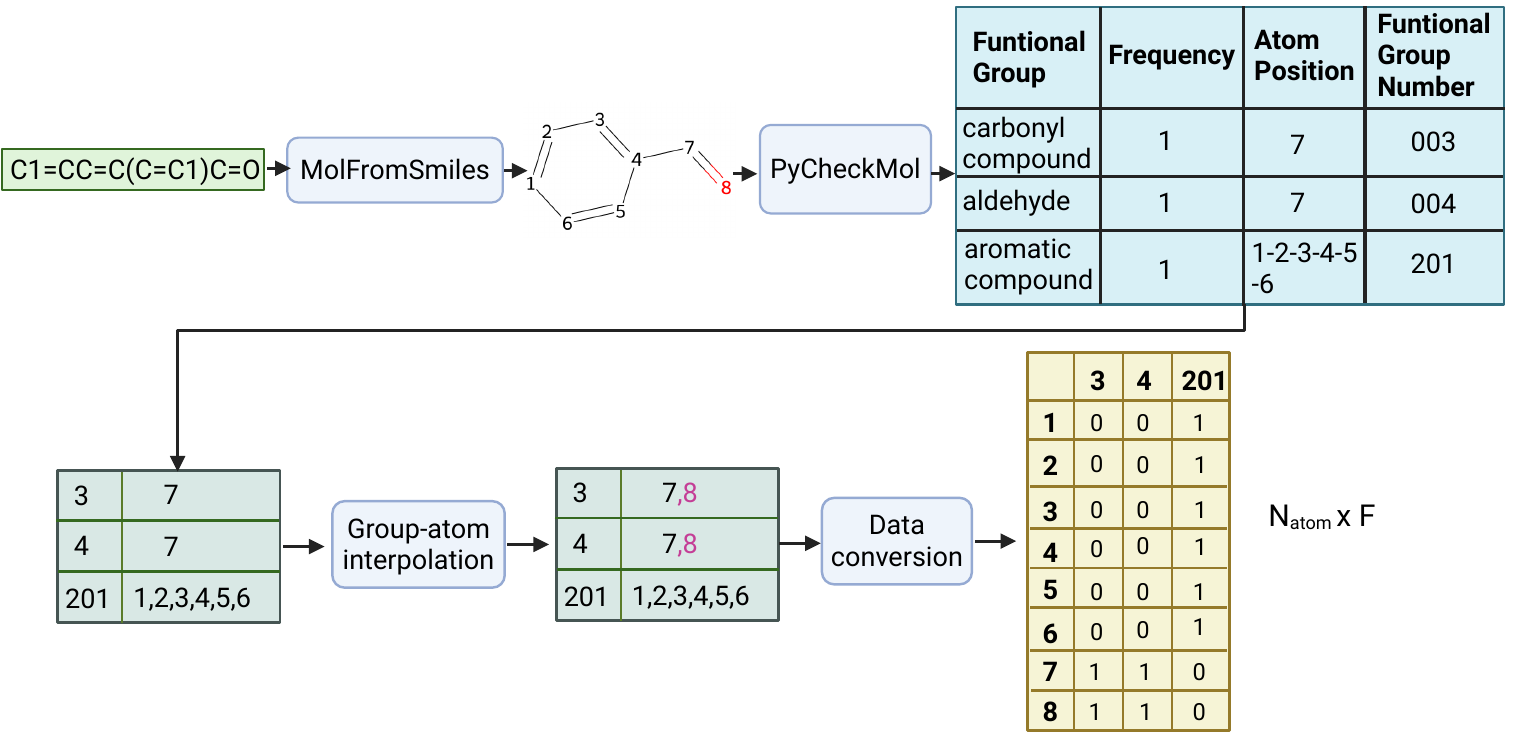}
\caption{Overview of the FGParser module. Given a ligand SMILES, RDKit is used to construct a molecular graph, which is then analyzed by PyCheckmol to detect predefined functional groups and generate initial atom-group mappings. A group-atom interpolation step ensures that all atoms are assigned to at least one functional group. The final output is a binary atom-group matrix.}
\label{fig:fgparser}
\end{figure}

The FGParser module decomposes a ligand into its constituent functional groups and generates an atom-group representation matrix for downstream processing, as illustrated in Figure~\ref{fig:fgparser}. The procedure comprises two stages: atom-group mapping and group-atom interpolation.

\paragraph{Atom-group mapping.}
Given a ligand SMILES sequence, denoted $\mathbf{D}$, FGParser first uses RDKit \cite{landrum2016rdkit} to convert the SMILES into a Mol object, which encodes the atom and bond information in a structured format suitable for cheminformatics operations. The functional groups are then identified using PyCheckmol~\cite{pyCheckmol}, which applies a curated set of predefined chemical patterns to detect chemically significant substructures.
Each detected functional group is assigned a unique identifier and all atoms belonging to that group are tagged with the corresponding ID. This process produces an initial mapping of functional groups to their member atoms, preserving explicit atom-group associations for downstream processing.

\paragraph{Group-atom interpolation.}
Some atoms may not belong to any canonical functional group identified in the first stage. To ensure full coverage, these atoms are assigned to the nearest functional group based on distance to functional-group atoms. Ties between equidistant groups are resolved by the ascending group ID. 

After this assignment step, we construct a binary matrix: 
$\mathbf{M} \in \mathbb{R}^{N_{\text{atom}} \times F}$ defined as
\[
\mathbf{M}_{i,j} =
\begin{cases}
1 & \text{if atom $i$ belongs to or is assigned to functional group $j$}, \\
0 & \text{otherwise}.
\end{cases}
\]
This conversion preserves atom-level resolution while incorporating functional group membership, enabling downstream modules to utilize both structural and chemical information.

\subsection{Functional Group and Positional Embedding for Molecular Identification (FINGER-ID)} \label{sec:fingerid}

\begin{figure}[t]
    \centering
    \includegraphics[width=\linewidth]{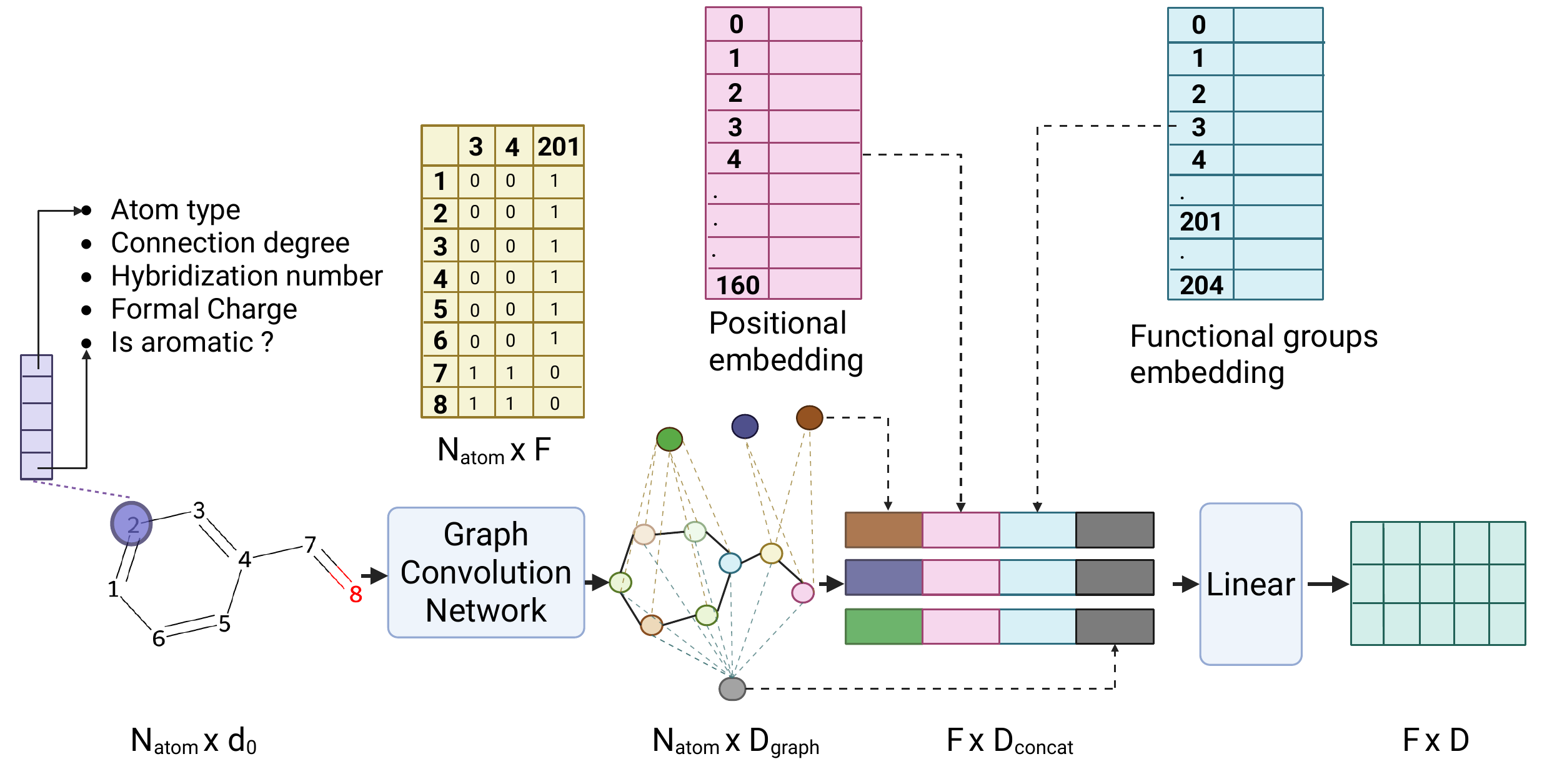}
    \caption{
        Atom-level features and chemical bonds define the molecular graph, which is encoded by a graph convolutional network (GCN). 
        The resulting atom embeddings are aggregated into functional group representations and enriched with group-type, positional, and global information. 
        These enriched embeddings are then projected to form the final ligand representation. 
        This hierarchical design captures structural signals from atomic, meso, intermediate, and global scales.}
    \label{fig:fingerid}
\end{figure}

The FINGER-ID module generates hierarchical context-aware embeddings for functional groups identified by FGParser. 
Unlike conventional fingerprints such as Morgan fingerprints, which cannot capture positional information, distinguish multiple occurrences of the same group, or incorporate explicit structural context, 
FINGER-ID integrates signals across the atomic, meso, intermediate, and global scales. 
This design yields chemically interpretable representations suitable for various downstream tasks (Figure~\ref{fig:fingerid}).

\paragraph{Atomic-scale (Atomic Properties) Initialization.}
We represent a molecule as a graph \(G=(V,E)\) with \(|V|=N\) atoms. 
Each atom \(a \in V\) is associated with a feature vector \(\mathbf{x}_a \in \mathbb{R}^{d_0}\) that encodes its atomic properties. 
In our implementation, we set \(d_0=5\), which corresponds to the type, degree, hybridization, formal charge, and aromaticity of the atom. 
The node features are collected into a matrix:
\[
X = [\mathbf{x}_1^\top; \dots; \mathbf{x}_N^\top] \in \mathbb{R}^{N \times d_0}.
\]
The edge set \(E\) is defined by chemical bonds that capture molecular connectivity.

\paragraph{Meso-scale (Neighborhood) Encoding.}
We encode the graph topology and node features with a graph convolutional network (GCN) to obtain latent atom embeddings:
\[
\mathbf{Z} = \mathrm{GCN}(G, X) \in \mathbb{R}^{N \times D_{\text{graph}}},
\]
where the $a$-th row $\mathbf{z}_a$ represents the embedding of atom $a$. 
This stage captures meso-scale connectivity patterns within atomic neighborhoods based solely on adjacency information.

\paragraph{Intermediate-scale (Functional Group) Embedding.}
Each functional group $g$ is represented by aggregating the embeddings of its constituent atoms according to the atom-to-group mapping $\mathcal{M}(g) \subseteq V$ obtained from FGParser:
\[
\mathbf{z}^{\text{inter}}_g = \mathrm{AGG}\{\mathbf{z}_a \mid a \in \mathcal{M}(g)\}, \quad g = 1,\dots,F,
\]
where $\mathrm{AGG}$ denotes mean pooling. 
The group-level representation is then concatenated with a learnable group-type embedding $\mathbf{e}^{\text{FG}}_g$ and a positional embedding $\mathbf{e}^{\text{pos}}_g$ to distinguish multiple occurrences of the same group.

\paragraph{Global-scale (Molecular) Embedding.}
A global embedding is derived by applying a READOUT operation over all atom embeddings:
\[
\mathbf{z}^{\text{global}} = \mathrm{READOUT}(\mathbf{Z}).
\]
Finally, each functional group embedding is augmented with a global molecular context.
\[
\mathbf{h}_g = \big[ \mathbf{z}^{\text{inter}}_g \;\Vert\; \mathbf{e}^{\text{FG}}_g \;\Vert\; \mathbf{e}^{\text{pos}}_g \;\Vert\; \mathbf{z}^{\text{global}} \big],
\]
yielding a multiscale representation that integrates atomic features, meso-scale neighborhoods, functional-group information, and global molecular structure.

\paragraph{Final Projection.}
All enriched group embeddings are stacked and linearly projected to obtain the final ligand representation:
\[
\mathbf{H} = [\mathbf{h}_1; \dots; \mathbf{h}_F] \in \mathbb{R}^{F \times D_{\text{concat}}}, \quad
\mathbf{H}_l = \mathbf{H} \mathbf{W} + \mathbf{b} \in \mathbb{R}^{F \times D}.
\]

By integrating atomic, meso, intermediate, and global scale information, 
FINGER-ID produces position-aware, chemically interpretable embeddings that can differentiate identical functional groups in distinct chemical environments, thereby enhancing downstream binding and molecular-property prediction.

\subsection{Self \& Cross Attention Transformer (SCAT)}
\begin{figure}[h]
    \centering
    \includegraphics[width=0.8\linewidth]{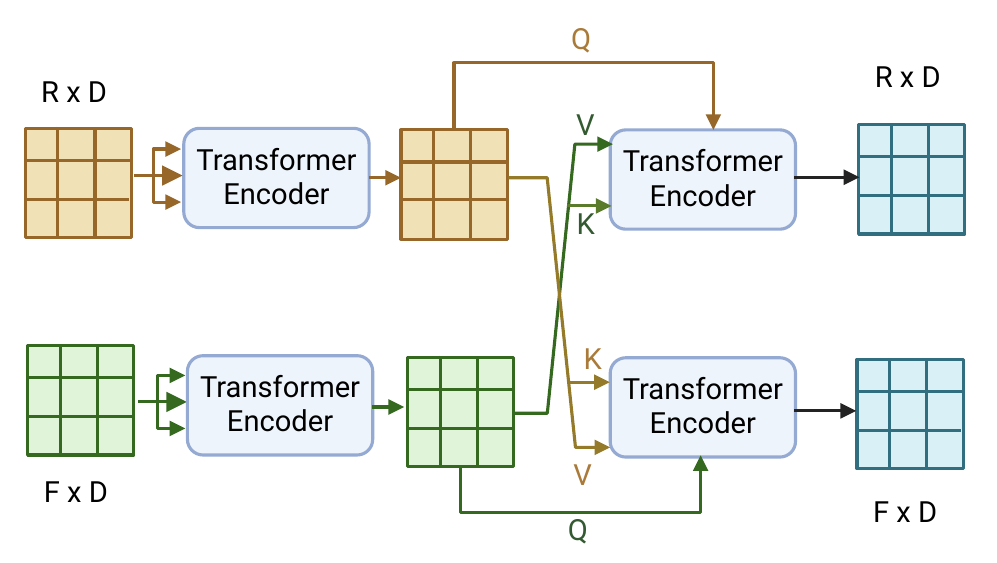}
    \caption{
    Overview of the Self \& Cross Attention Transformer (SCAT) module, which integrates intra-molecular self-attention and inter-molecular cross-attention between protein residues and ligand functional groups.}
    \label{fig:scat_arch}
\end{figure}

To capture both intra- and inter-molecular dependencies, we introduce the SCAT module, as illustrated in Figure~\ref{fig:scat_arch}. Let
$\mathbf{H}_p \in \mathbb{R}^{R \times D}$ denote the protein residue embeddings obtained from ESM C, and
$\mathbf{H}_l \in \mathbb{R}^{F \times D}$ denote the functional group embeddings of the ligand produced by FINGER-ID.

\paragraph{Self-attention.} We utilize the self-attention mechanism to further encode the residue and functional-group embeddings:
\begin{align*}
\mathbf{H}'_p &= \text{SA}_p(\mathbf{H}_p), \\
\mathbf{H}'_l &= \text{SA}_l(\mathbf{H}_l),
\end{align*}
where $\text{SA}_p$ and $\text{SA}_l$ are transformer encoder blocks with self-attention applied within each modality.

\paragraph{Cross-attention.} Furthermore, we use the cross-attention to encode the outputs from the self-attention:
\begin{align*}
\mathbf{H}''_p &= \text{CA}_{p \leftarrow l}(\mathbf{H}'_p, \mathbf{H}'_l), \\
\mathbf{H}''_l &= \text{CA}_{l \leftarrow p}(\mathbf{H}'_l, \mathbf{H}'_p),
\end{align*}
where $\text{CA}_{p \leftarrow l}$ attends from protein residues (queries) to ligand functional groups (keys/values), and $\text{CA}_{l \leftarrow p}$ performs the reverse.

The final outputs $\mathbf{H}''_p$ and $\mathbf{H}''_l$ are enriched with local context and cross-molecular interaction signals, providing chemically and biologically informed representations for downstream interaction prediction.

\subsection{PairwiseUNet}
\begin{figure}[ht]
    \centering
    \includegraphics[width=\linewidth]{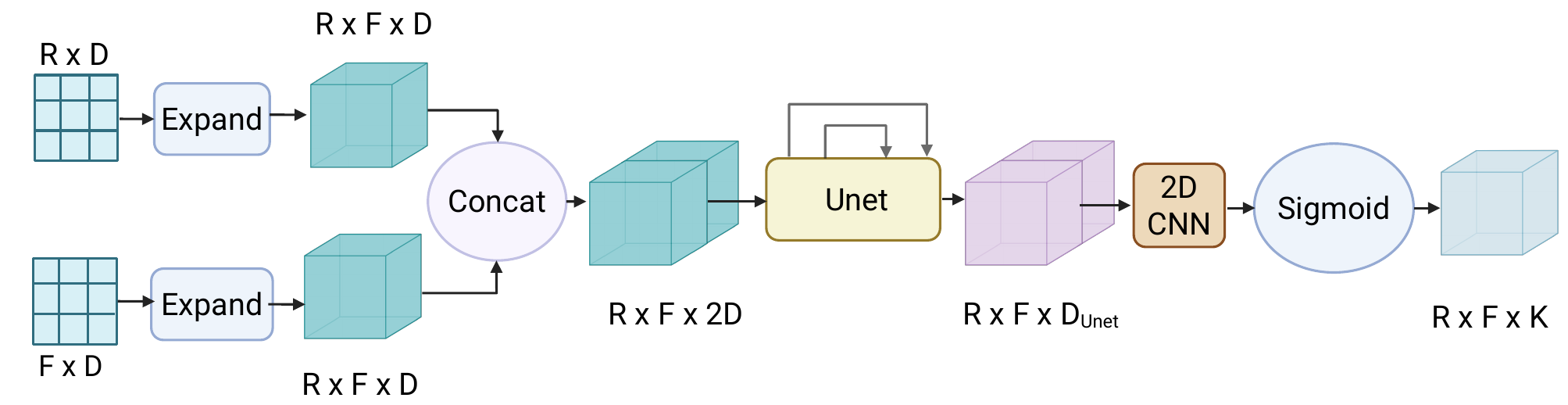}
    \caption{
    Overview of the PairwiseUNet module, which constructs residue–functional group pairwise tensors and processes them with a U-Net–inspired 2D CNN architecture, followed by a sigmoid layer, to predict interaction probability maps.}
    \label{fig:pairwise_unet}
\end{figure}

The PairwiseUNet module, as illustrated in Figure~\ref{fig:pairwise_unet}, takes the enriched protein and ligand representations from SCAT, 
$\mathbf{H}''_p \in \mathbb{R}^{R \times D}$ and $\mathbf{H}''_l \in \mathbb{R}^{F \times D}$, and transforms them into a structured pairwise feature tensor for interaction prediction.

\paragraph{Pairwise tensor construction.}
Protein embeddings are transmitted along the functional group dimension to form
$\mathbf{E}_p \in \mathbb{R}^{R \times F \times D}$, while ligand embeddings are transmitted along the residue dimension to form
$\mathbf{E}_l \in \mathbb{R}^{R \times F \times D}$. 
These tensors are concatenated along the feature dimension, producing a joint representation:
\begin{align*}
\mathbf{Z} &= \text{Concat}(\mathbf{E}_p, \mathbf{E}_l) \in \mathbb{R}^{R \times F \times 2D}.
\end{align*}

\paragraph{2D U-Net processing.}
The combined tensor $\mathbf{Z}$ is processed by a 2D U-Net, which models both local and long-range spatial dependencies across the residue--functional group grid:
\begin{align*}
\mathbf{U} &= \text{UNet}(\mathbf{Z}) \in \mathbb{R}^{R \times F \times D_{\text{unet}}}.
\end{align*}

\paragraph{Interaction type prediction.}
Finally, a stack of 2D convolutional layers~\cite{lecun2002gradient} maps the pairwise feature tensor $\mathbf{U}$ to interaction-type probabilities:
\begin{align*}
\mathbf{P} &= \text{Sigmoid}(\text{Conv2D}(\mathbf{U})) \in [0,1]^{R \times F \times K},
\end{align*}
where $K=7$ corresponds to biologically defined interaction types (e.g. hydrogen bonds, $\pi$–stacking, salt bridges). 
This produces an interpretable interaction map that can be directly derived from sequence-level input, with each entry $\mathbf{P}_{r,f,k}$ representing the independent probability of residue $r$ and functional group $f$ participating in the interaction type $k$.

\subsection{Binding Affinity Predictor}
\label{app:affinity_predictor}
\begin{figure}[t]
    \centering
    \includegraphics[width=\linewidth]{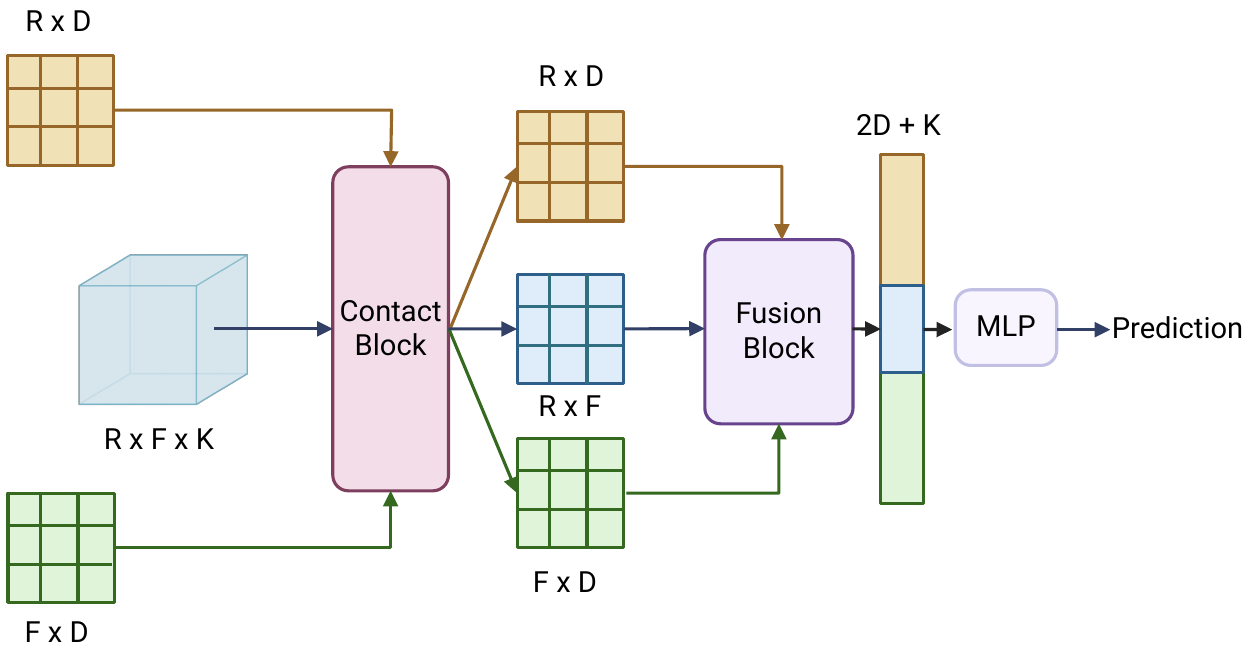}
    \caption{
    Overview of the binding affinity prediction framework. 
    Target embeddings $\mathbf{H}_p$ from the protein language model and ligand embeddings $\mathbf{H}_l$ from the FINGER-ID module are combined with interaction probabilities $\mathbf{P}$ from the PairwiseUnet module. 
    The \textbf{ContactBlock} encodes residue-functional group interactions via edge strengths, bidirectional attention, and context aggregation, producing enriched embeddings. 
    The \textbf{FusionBlock} pools these embeddings using learned importance weights and incorporates global interaction type contributions to form the final feature vector $\mathbf{h}$, which is fed into an MLP to predict the binding affinity.
    }
    \label{fig:binding_affinity_prediction}
\end{figure}

All weights of the modules in Figure~\ref{fig:framework} are frozen after training the interaction prediction module. 
Given target embeddings
\(\mathbf{H}_p \in \mathbb{R}^{R \times D}\) from the Protein Language Model (ESM-C) and ligand embeddings
\(\mathbf{H}_l \in \mathbb{R}^{F \times D}\) from FINGER-ID, along with interaction probabilities
\(\mathbf{P} \in \mathbb{R}^{R \times F \times K}\) from PairwiseUnet, the framework calculates a scalar prediction per protein-ligand complex (Figure~\ref{fig:binding_affinity_prediction}).

\paragraph{Contact Block.} 
This block fuses residue and functional group information using predicted pairwise interaction probabilities. It first computes an edge-strength matrix that quantifies the importance of each residue-functional-group pair. From these strengths, it derives bidirectional attention coefficients and aggregates neighbor information into context vectors, which are then projected and added to the original embeddings to form enriched representations.
\begin{itemize}
\item \textbf{Edge strength.} Edge strengths aggregate contributions from all interaction types into a single scalar per residue-functional-group pair:
\begin{align*}
    S_{r,f} &= \sum_{k=1}^{K} P_{r,f,k}\, w_k, \quad
    S = [S_{r,f}] \in \mathbb{R}^{R\times F},
\end{align*}
where $w_k\in\mathbb{R}$ are learnable importance weights that reweight the predicted interaction-type probabilities.

\item \textbf{Bidirectional attention.} We convert edge strengths to normalized attention coefficients to capture the asymmetric relevance of each node to its neighbors:
\begin{align*}
    \alpha^{p\to l}_{r,f} &= \frac{\exp(S_{r,f})}{\sum_{f'=1}^{F}\exp(S_{r,f'})} \in \mathbb{R}, \quad
    \alpha^{l\to p}_{r,f} = \frac{\exp(S_{r,f})}{\sum_{r'=1}^{R}\exp(S_{r',f})} \in \mathbb{R}.
\end{align*}
where $\alpha^{r\to f}_{r,f}$ measures the importance of the functional group $f$ for residue $r$, and $\alpha^{f\to r}_{r,f}$ the opposite.

\item \textbf{Context aggregation \& enriched embeddings.} Each residue aggregates a weighted sum of its functional-group neighbors to form a context vector, and each functional group aggregates a weighted sum of its residue neighbors to form its context vector. These context vectors are then projected and added to the original embeddings to produce enriched representations:
\begin{align*}
    \mathbf{c}^{l}_{r} &= \sum_{f=1}^{F} \alpha^{p\to l}_{r,f}\,\mathbf{H}_l[f] \in \mathbb{R}^{D}, \quad
    \mathbf{c}^{p}_{f} = \sum_{r=1}^{R} \alpha^{l\to p}_{r,f}\,\mathbf{H}_p[r] \in \mathbb{R}^{D},\\
    \mathbf{H}'_p[r] &= \mathbf{H}_p[r] + \mathrm{proj}_p(\mathbf{c}^{l}_{r}) \in \mathbb{R}^{D}, \quad
    \mathbf{H}'_l[f] = \mathbf{H}_l[f] + \mathrm{proj}_l(\mathbf{c}^{p}_{f}) \in \mathbb{R}^{D},\\
    \mathbf{H}'_p &= [\mathbf{H}'_p[r]]_{r=1}^R \in \mathbb{R}^{R \times D}, \quad 
    \mathbf{H}'_l = [\mathbf{H}'_l[f]]_{f=1}^F \in \mathbb{R}^{F \times D}.
\end{align*}
Here, $\mathrm{proj}_r(\cdot)$ and $\mathrm{proj}_f(\cdot)$ denote linear projections (learned) that map the context vectors back to the embedding space before residual addition.
\end{itemize}

\paragraph{Fusion Block.} 
This block summarizes the enriched embeddings into compact pooled vectors and encodes a global signal for each interaction type. The pooled vectors are constructed so that nodes with stronger overall edge connections receive greater importance.
\begin{itemize}
\item \textbf{Pooling weights and pooled embeddings.} We compute per-node importance scores by summing incident edge strengths, normalize them to obtain pooling weights, and form pooled embeddings as weighted sums of enriched node embeddings:
\begin{align*}
    s^{p}_r &= \sum_{f=1}^{F} S_{r,f} \in \mathbb{R}, \quad 
    s^{l}_f = \sum_{r=1}^{R} S_{r,f} \in \mathbb{R},\\
    \beta^{p}_r &= \frac{s^p_r}{\sum_{r'} s^p_{r'}} \in \mathbb{R}, \quad
    \beta^{l}_f = \frac{s^l_f}{\sum_{f'} s^l_{f'}} \in \mathbb{R},\\
    \mathbf{H}^{\text{pool}}_p &= \sum_{r=1}^{R} \beta^p_r\,\mathbf{H}'_p[r] \in \mathbb{R}^{D}, \quad
    \mathbf{H}^{\text{pool}}_l = \sum_{f=1}^{F} \beta^l_f\,\mathbf{H}'_l[f] \in \mathbb{R}^{D}.
\end{align*}
Intuitively, nodes with larger total edge strength contribute more to the pooled representation.

\item \textbf{Global interaction-type contribution.} A compact global vector records the overall contribution of each interaction type throughout the complex:
\begin{align*}
    s_k &= \sum_{r=1}^{R}\sum_{f=1}^{F} P_{r,f,k}\, w_k \in \mathbb{R}, \quad
    \mathbf{s} = [s_1,\ldots,s_K]^\top \in \mathbb{R}^{K}.
\end{align*}
which captures the aggregate presence and importance of the interaction type $k$ in the current complex.

\end{itemize}

\paragraph{Final representation and prediction.}
We obtain the final feature vector by concatenating the pooled target embedding, the global interaction-type vector, and the pooled ligand embedding and then pass it to an MLP to produce the scalar binding affinity prediction.
\begin{align*}
\mathbf{h} &= \big[\,\mathbf{H}^{\text{pool}}_p,\; \mathbf{s},\; 
    \mathbf{H}^{\text{pool}}_l\,\big] \in \mathbb{R}^{2D+K},\\
\hat{y} &= \mathrm{MLP}(\mathbf{h}) \in \mathbb{R}.
\end{align*}

\section{Implementation Details}
\label{sec:appendix_implementation_details}

\subsection{Training Procedure}

We trained our model on the LP-PDBBind dataset, which contains protein-ligand complexes with annotated 3D structures and experimentally measured binding affinities. Residue-functional group interaction labels were derived using PLIP for geometric annotation and PyCheckMol for chemical functional group decomposition.

\paragraph{LINKER training.} To address the extreme class imbalance inherent in interaction annotations, we adopt the Focal Loss~\cite{lin2017focal}, which dynamically weights down-weights well-classified examples and emphasizes harder misclassified instances. The Focal Loss is defined as:
\begin{align*}
\mathcal{L}_{\text{focal}} = - \alpha (1 - p_t)^\gamma \log(p_t),
\end{align*}
where $p_t$ is the predicted probability for the true class, $\alpha$ balances the contribution of positive and negative examples, and $\gamma$ controls the degree of focus on hard examples. In our experiments, we set $\alpha = 0.85$ and $\gamma = 1.0$.

\paragraph{Binding Affinity Predictor training.}  
For downstream regression, we introduce an interaction-aware neural network (see Appendix~\ref{app:affinity_predictor}) that integrates residue embeddings, functional group embeddings, and interaction probabilities produced by LINKER. To train the binding affinity predictor while regularizing the latent space, we optimize a combination of the Mean Squared Error (MSE) loss and the latent alignment loss:
\begin{align*}
\mathcal{L}_{\text{total}} &= \mathcal{L}_{\text{MSE}} + \beta \, \mathcal{L}_{\text{latent}}, 
\end{align*}
where $\beta$ is a hyperparameter controlling the contribution of the latent alignment term.  

The MSE loss measures the difference between predicted and true binding affinities:
\begin{align*}
\mathcal{L}_{\text{MSE}} &= \frac{1}{N} \sum_{i=1}^{N} (y_i - \hat{y}_i)^2,
\end{align*}
where $y_i$ and $\hat{y}_i$ denote the ground truth and predicted binding affinities, respectively.

The latent alignment loss combines an InfoNCE term \cite{oord2018representation} and a uniformity term \cite{wang2020understanding}:
\begin{align*}
\mathcal{L}_{\text{latent}} &= \mathcal{L}_{\text{InfoNCE}} + \lambda \, \mathcal{L}_{\text{uniform}},
\end{align*}
where $\lambda$ balances the contribution of the uniformity term.

The latent alignment loss encourages embeddings of samples with similar binding affinities to be close in the latent space while maintaining uniformity in the hypersphere. Let $\mathbf{h}$ denote the final representation defined in Appendix~\ref{app:affinity_predictor}, and let $\mathbf{h}_i$ be the representation of the $i$-th sample in a batch. We first normalize each sample embedding to obtain:
\begin{align*}
\mathbf{z}_i = \frac{\mathbf{h}_i}{\|\mathbf{h}_i\|_2} \in \mathbb{R}^{2D+K}.
\end{align*}
which projects all embeddings onto the unit hypersphere. This normalization removes the influence of vector magnitude and ensures that similarity is determined solely by the angular distance, thereby preventing trivial solutions where embeddings collapse or scale arbitrarily.

The InfoNCE loss encourages embeddings of samples with similar binding affinities to be close to each other in the latent space. For each sample $i$, we define its positive sample $p(i)$ as the sample in the batch with the most similar binding score (excluding itself). The loss is then computed as:
\begin{align*}
\mathcal{L}_{\text{InfoNCE}} &= - \frac{1}{B} \sum_{i=1}^{B} 
\log \frac{\exp\big( \mathbf{z}_i^\top \mathbf{z}_{p(i)} / \tau \big)}
{\sum_{j=1}^{B} \exp\big( \mathbf{z}_i^\top \mathbf{z}_j / \tau \big)},
\end{align*}
where $\tau$ is a temperature hyperparameter that controls the sharpness of the similarity distribution.

The Uniformity loss encourages latent vectors to spread on the hypersphere:
\begin{align*}
\mathcal{L}_{\text{uniform}} &= \log \frac{1}{B^2} \sum_{i,j=1}^{B} \exp \Big( - 2 \| \mathbf{z}_i - \mathbf{z}_j \|_2^2 \Big).
\end{align*}

In our experiments, we set $\beta = 2$, $\lambda = 0.1$, and $\tau = 0.1$.

\paragraph{Optimization Setup.} 
LINKER was trained for 30 epochs and the Binding Affinity Predictor for 80 epochs, both using the Adam optimizer with a learning rate of $2 \times 10^{-5}$ and batch sizes of 2 and 16, respectively. Validation performance was monitored after each epoch to assess generalization.

\paragraph{Implementation Environment.} 
All experiments were implemented in PyTorch~(v2.6.0) with CUDA~12.4 and run on an NVIDIA Tesla P100 GPU (16 GB VRAM).

\subsection{Evaluation Metric}

We use the following evaluation metrics in our experiments:

\paragraph{Precision-Recall Curve (PR Curve).}
The PR curve evaluates the trade-off between precision and recall across varying thresholds:
\[
\text{Precision} = \frac{\text{TP}}{\text{TP} + \text{FP}}, \qquad
\text{Recall} = \frac{\text{TP}}{\text{TP} + \text{FN}}.
\]
The area under the PR curve (PR AUC) summarizes model performance under class imbalance, where positive labels are sparse.

\paragraph{Receiver Operating Characteristic Curve (ROC Curve).}
The ROC curve plots the true positive rate (TPR) against the false positive rate (FPR):
\[
\text{TPR} = \frac{\text{TP}}{\text{TP} + \text{FN}}, \qquad
\text{FPR} = \frac{\text{FP}}{\text{FP} + \text{TN}}.
\]
The area under the ROC curve (ROC AUC) reflects the model's ability to discriminate between positive and negative classes.

\paragraph{Weighted Precision.}
For soft labels, we compute weighted precision by thresholding the model's attention scores and weighting predictions by the smoothed label values:
\[
\text{Weighted Precision} = \frac{\sum_{i=1}^{N} \hat{y}_i \cdot y_i}{\sum_{i=1}^{N} \hat{y}_i},
\]
where \( \hat{y}_i \in \{0, 1\} \) is the binary prediction and \( y_i \in [0, 1] \) is the smoothed supervision label.
\paragraph{Prevalence.}
Prevalence is simply the proportion of positive samples in the dataset:
\[
\text{Prevalence} = \frac{1}{N}\sum_{i=1}^{N} y_i,
\]
which corresponds to the horizontal baseline of random predictions in the PR curve.

\paragraph{Enrichment.}
Enrichment measures the improvement of the precision of a model over the prevalence baseline:
\[
\text{Enrichment} = \frac{\text{Precision}}{\text{Prevalence}}.
\]
High enrichment values indicate that the model retrieves substantially more true positives than expected under random selection, which is particularly informative at low recall in highly imbalanced settings.

\paragraph{Root Mean Squared Error (RMSE).}
For regression tasks, we use the RMSE to quantify the deviation between predicted and ground-truth values:
\[
\text{RMSE} = \sqrt{\frac{1}{N} \sum_{i=1}^{N} \left( \hat{y}_i - y_i \right)^2 },
\]
where \( \hat{y}_i \) denotes the predicted binding affinity and \( y_i \) is the corresponding ground-truth label.
Lower RMSE values indicate better predictive accuracy.

\end{document}